\documentclass{aifrontiers}
\usepackage{aifrontiers}
\usepackage{booktabs}
\usepackage{array}
\usepackage{colortbl}
\usepackage{caption}
\usepackage{subcaption}
\usepackage{multirow}
\usepackage{float}
\usepackage{enumitem}
\usepackage{xspace}
\usepackage{makecell}  

\usepackage[T1]{fontenc}    
\usepackage[utf8]{inputenc} 

\newcommand{\tabref}[1]{Table~\ref{#1}}
\newcommand{\figref}[1]{Figure~\ref{#1}}
\newcommand{\secref}[1]{Section~\ref{#1}}

\newlength\savewidth
\newcommand{\tablestyle}[2]{\setlength{\tabcolsep}{#1}\renewcommand{\arraystretch}{#2}\centering\footnotesize}

\newcolumntype{x}[1]{>{\centering\arraybackslash}p{#1pt}}
\newcolumntype{y}[1]{>{\raggedright\arraybackslash}p{#1pt}}

\definecolor{baselinecolor}{gray}{0.93}

\newcommand{\phivr}{Phi-4-reasoning-vision-15B\xspace}

\begin{document}

\title{\phivr Technical Report}
\shorttitle{\phivr}
\author{
     Jyoti Aneja, Michael Harrison, Neel Joshi,\\Tyler LaBonte, John Langford, Eduardo Salinas\\[4pt]
    {\normalsize Microsoft Research}
}
\date{\today}

\renewcommand{\weblink}{https://aka.ms/Phi-4-reasoning-vision}
\renewcommand{\foundrylink}{https://aka.ms/Phi-4-r-v-FoundryLabs}
\renewcommand{\hflink}{https://huggingface.co/microsoft/Phi-4-reasoning-vision-15B}
\renewcommand{\ghlink}{https://github.com/microsoft/Phi-4-reasoning-vision-15B}

\begin{abstract}
We present \phivr, a compact open-weight multimodal reasoning model, and share the motivations, design choices, experiments, and learnings that informed its development. Our goal is to contribute practical insight to the research community on building smaller, efficient multimodal reasoning models and to share the result of these learnings as an open-weight model that is good at common vision and language tasks and excels at scientific and mathematical reasoning and  understanding user
interfaces. Our contributions include demonstrating that careful architecture choices and rigorous data curation enable smaller, open-weight multimodal models to achieve competitive performance with significantly less training and inference-time compute and tokens. The most substantial improvements come from systematic filtering, error correction, and synthetic augmentation—reinforcing that data quality remains the primary lever for model performance. Systematic ablations show that high-resolution, dynamic-resolution encoders
yield consistent improvements, as accurate perception is a prerequisite for high-quality reasoning. Finally, a hybrid mix of reasoning and non-reasoning data with explicit mode tokens allows a single model to deliver fast direct answers for simpler tasks and chain-of-thought reasoning for complex problems.

\end{abstract}

\maketitle


\section{Introduction}
\label{sec:introduction}

\begin{figure}[t!]
    \centering
    \includegraphics[width=\linewidth]{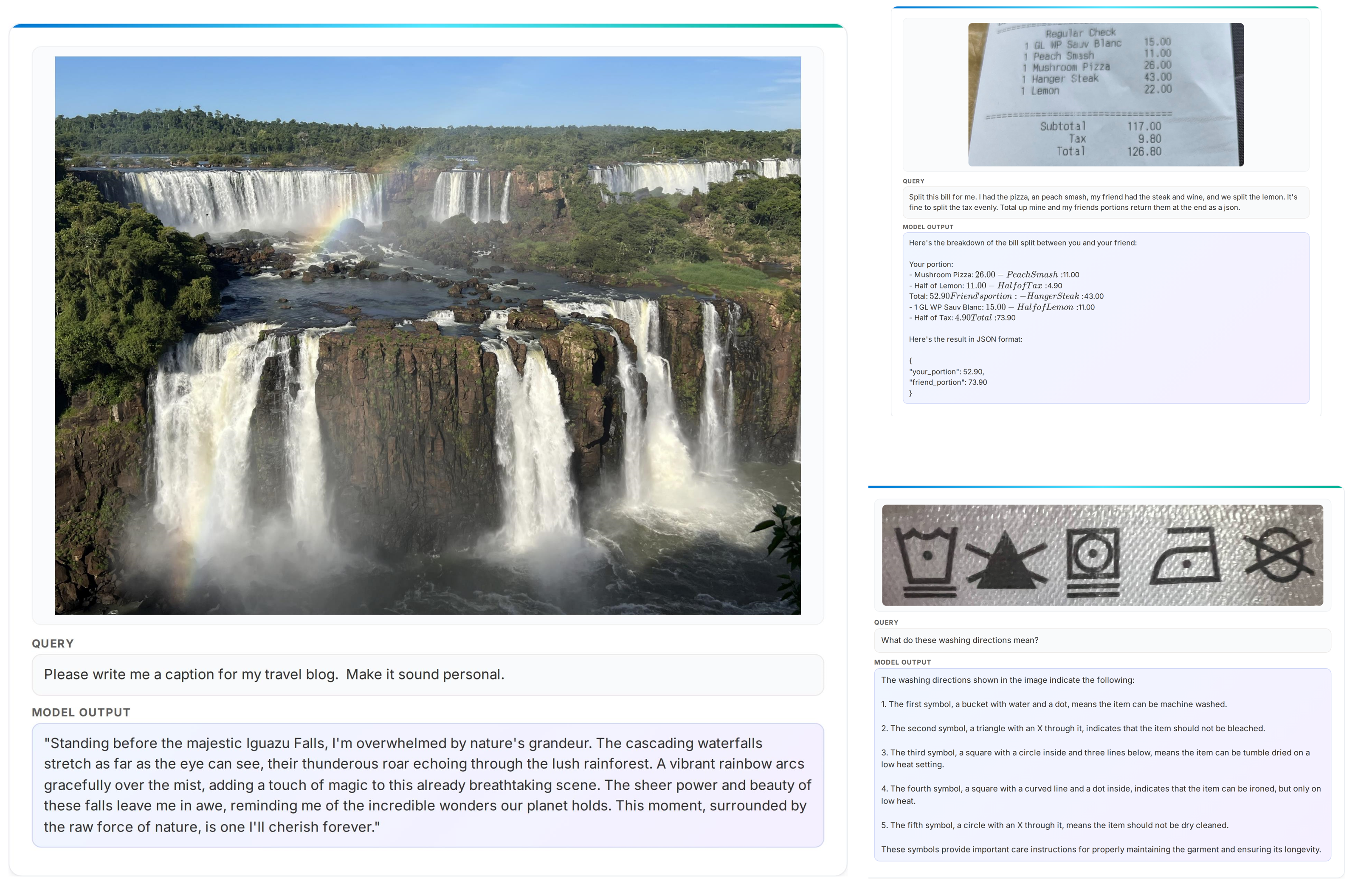}
    \caption{\phivr can help with a wide range of everyday tasks, from writing travel captions and interpreting receipts to reading garment care instructions.}
    \label{fig:everyday}
\end{figure}

\begin{figure}[t]
    \centering
    \includegraphics[width=0.99\linewidth]{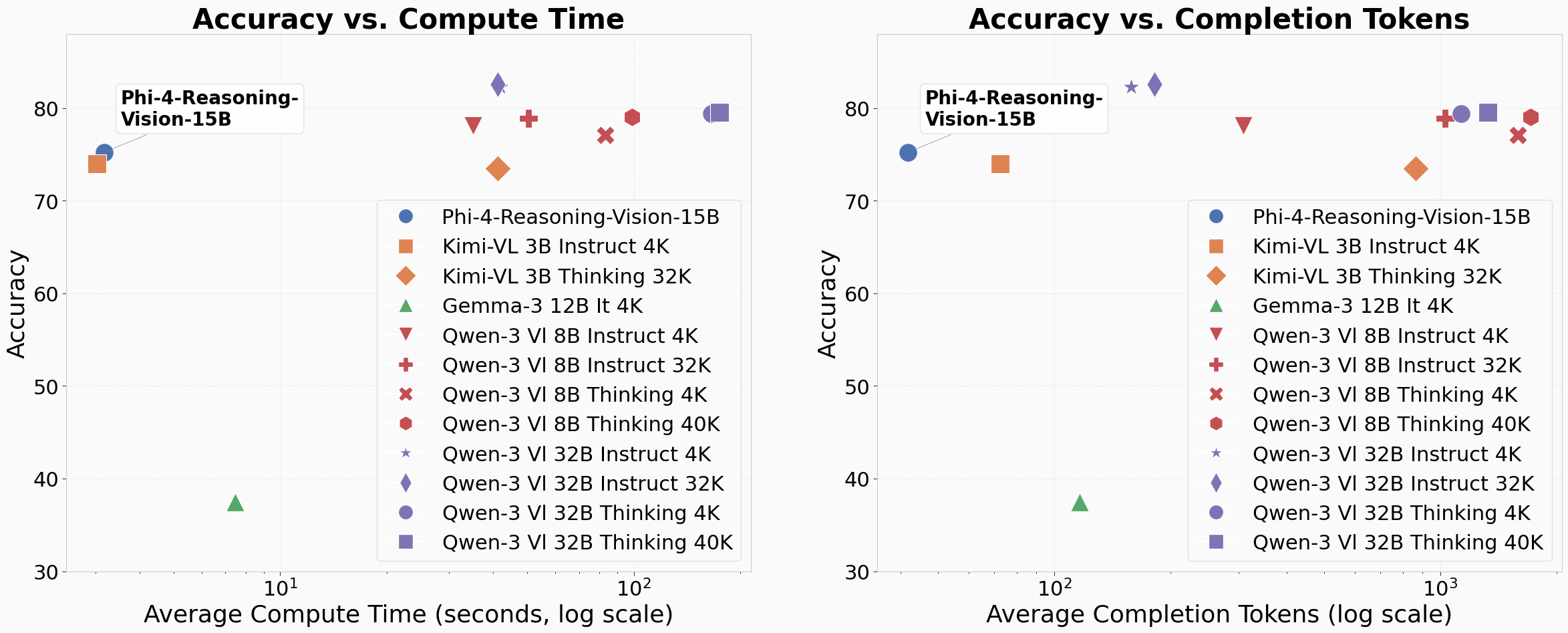}
    \caption{\phivr presents a compelling option compared to existing models, pushing the Pareto frontier of the trade-off between accuracy and compute costs. We achieve competitive performance with much slower models that require more time and tokens, and higher accuracy than similarly fast models. These values were computed by averaging accuracy, time, and output token counts for a subset of 4 benchmarks: ChartQA\textsubscript{TEST}, MathVista\textsubscript{MINI}, MMMU\textsubscript{VAL}, and ScreenSpot\_v2.}
    \label{fig:timing-and-tokens}
\end{figure}

\phivr is a compact open-weight multimodal reasoning model that balances reasoning power, efficiency, and training data needs. It is a broadly capable model that allows for natural interaction for a wide array of vision-language tasks and excels at math and science reasoning and understanding user interfaces, as shown in Figure~\ref{fig:everyday}.  Beyond these general capabilities, our model presents an appealing value relative to current open-weight models, pushing the Pareto frontier of the trade-off between accuracy and compute costs. We achieve competitive accuracy with much slower models that require ten times or more compute time and tokens, and better accuracy than similarly fast models, particularly when it comes to math and science reasoning, as shown in Figure~\ref{fig:timing-and-tokens}.

In this report, we share the motivations, design choices, experiments, and learnings that informed its development, as well as an evaluation of the model's performance and guidance on how to use it. Our goal is to contribute practical insight to the community on building smaller, efficient multimodal reasoning models and to share an open-weight model that is competitive with models of similar size at general vision-language tasks, excels at computer use, and at scientific and mathematical multimodal reasoning.

\subsection{Focus on Smaller and Faster Vision--Language Models}

Many popular vision--language models (VLMs) have trended towards growing in parameter count and 
the number of tokens they consume and generate. This leads to increased training and inference-time cost and latency, impeding their usability for downstream deployment, especially in resource-constrained or interactive settings.

A growing countertrend towards smaller models aims to boost efficiency, enabled by careful model design and data curation---a goal pioneered by the Phi~\citep{phi} family of models and furthered by \phivr. We specifically build on learnings from the Phi-4~\citep{phi4} and Phi-4-Reasoning~\citep{phi4reasoning} language models and show how a multimodal model can be trained to cover a wide range of vision and language tasks without relying on extremely large training datasets, architectures, or excessive inference-time token generation. Our model is intended to be lightweight enough to run on modest hardware while remaining capable of structured reasoning when it is beneficial. Our model was trained with far less compute than many recent open-weight VLMs of similar size. We used just 200 billion tokens of multimodal data leveraging Phi-4-Reasoning (trained with 16 billion tokens) based on a core model Phi-4 (400 billion unique tokens), compared to more than 1 trillion tokens used for training multimodal models like Qwen 3 VL~\citep{qwen3vl}, Kimi-VL~\citep{kimivl}, and Gemma3~\citep{gemma3}. We therefore present a compelling option compared to existing models, pushing the Pareto frontier of the trade-off between accuracy and compute costs.

\section{Architecture and Training}
\label{sec:architecture}

Training a multimodal reasoning model raises numerous questions and design decisions around model architecture, dataset quality and composition, training curriculum, and the interaction between reasoning-heavy and non-reasoning perception-focused tasks. All of these choices affect the learned behavior.

\subsection{Early vs.\ Mid Fusion}
\label{sec:fusion}

There are several options for model architecture based on when and how visual and textual information is fused. In late or mid-fusion models, a vision encoder first converts images into a compact set of visual tokens via a pretrained image encoder, which are then projected into the language embedding space and injected into a pretrained LLM~\citep{llava}. This approach enables meaningful cross-modal reasoning while preserving the strengths and scalability of large unimodal models. This approach keeps training and inference costs manageable, as it can utilize the power of pretrained components that have typically been trained on trillions of tokens.

Early-fusion models, by contrast, process all image patches and text tokens into a single transformer, allowing unrestricted cross-attention across modalities throughout the network~\citep{chameleonteam2025chameleonmixedmodalearlyfusionfoundation}. While this can yield richer joint representations and tighter visual--textual grounding, it significantly increases compute, memory, and data requirements. Given our goal of creating a highly performant model with less compute and data, we use a mid-fusion architecture. It offers a practical trade-off between expressivity and efficiency without the heavy cost of full early fusion.

\subsection{Vision Encoder and Image Processing}
\label{sec:vision-encoder}

\begin{figure}[t!]
    \centering
    \includegraphics[width=0.7\linewidth]{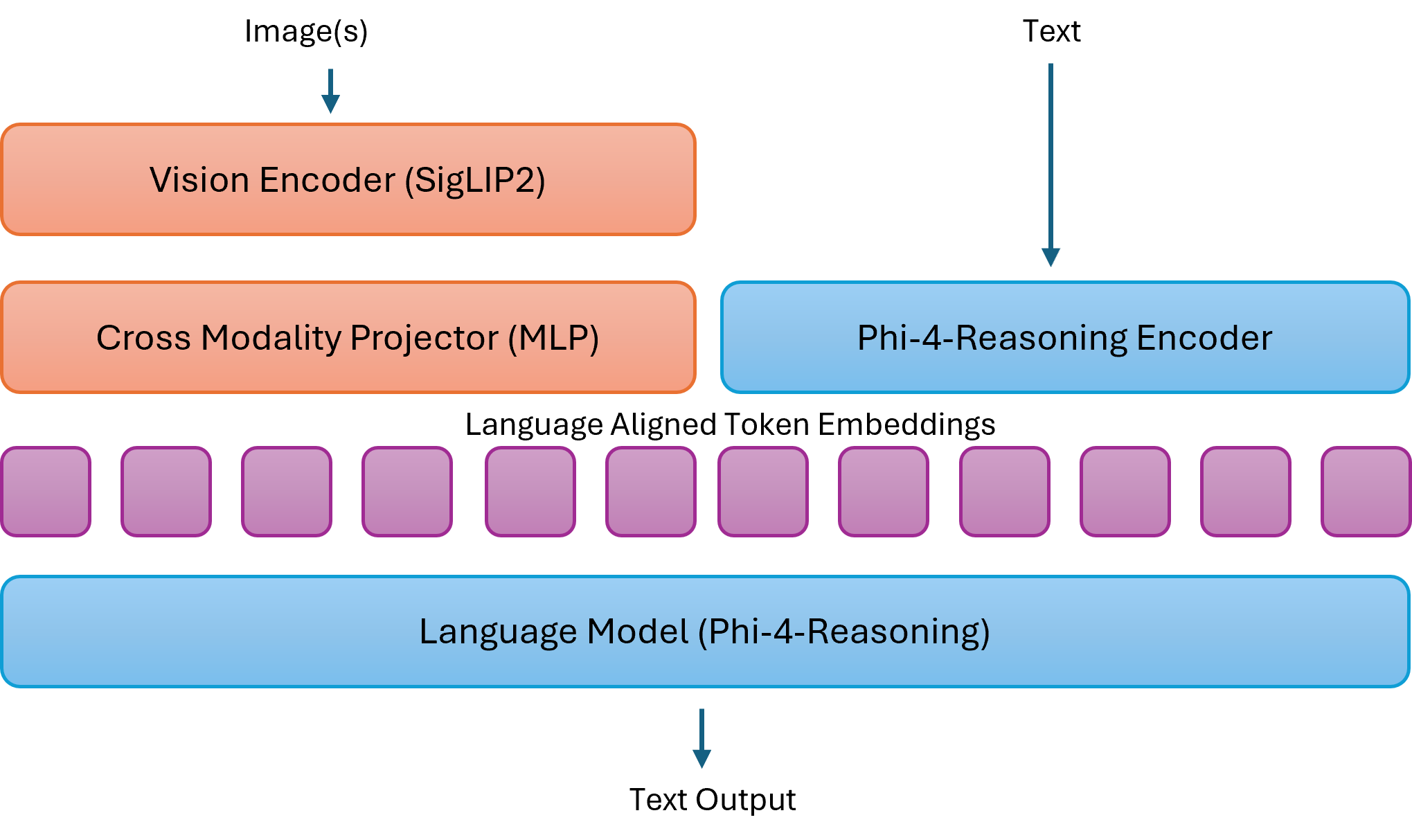}
    \caption{Overview of the \phivr mid-fusion architecture. Images are processed by a SigLIP-2 vision encoder and projected into the language embedding space via a cross-modality projector (MLP). The resulting visual ``soft'' tokens are interleaved with text tokens and fed into the Phi-4-Reasoning language model.}
    \label{fig:architecture}
\end{figure}

We build on the SigLIP-2 vision encoder~\citep{siglip2} and the Phi-4-Reasoning backbone, as shown in Figure~\ref{fig:architecture}. In our previous work, we found that multimodal language models sometimes struggled to solve tasks, not because of a lack of reasoning proficiency, but rather because of an inability to identify and extract relevant perceptual information from the image~\citep{balachandran2024eurekaevaluatingunderstandinglarge}. This problem compounds when considering computer-use and multimodal grounding tasks. In particular, desktop screens and browsers are information-dense with relatively small interactive elements, making fine-grained high-resolution feature extraction a prerequisite for agentic applications.

With high-resolution multimodal benchmarks increasing in relevance, several open-source multimodal language models have adapted their methodologies accordingly, e.g., Gemma3~\citep{gemma3} uses pan-and-scan, NVILA~\citep{nvila} uses dynamic S$^2$, and Qwen3-VL~\citep{qwen3vl} uses a bespoke vision encoder which operates at native resolution. However, their trade-offs are difficult to understand across different datasets and hyperparameters. To explore these options, we conducted a large-scale ablation of several vision encoder and image processing techniques, with the goal of understanding and maximizing grounding performance.

\begin{table}[b!]
\centering
\caption{Results with different resolution handling approaches on MathVista~\citep{mathvista}, ScreenSpot~\citep{seeclick}, ScreenSpot-Pro~\citep{screenspotpro}, and V*Bench~\citep{vstarbench}. We have \textbf{bolded} the top two configurations on each benchmark. These experiments are done on a smaller, 5B, variation of our model created for testing purposes.}
\label{tab:vision-encoder}
\tablestyle{5pt}{1.2}
\begin{tabular}{l c c c c c}
\toprule
\textbf{Method} & \textbf{Max Tokens} & \textbf{MathVista} & \textbf{ScreenSpot} & \textbf{ScreenSpot-Pro} & \textbf{V*Bench} \\
\midrule
Dynamic-S$^2$         & 3096 & 42.9          & 78.4          & 9.4           & 52.9 \\
Multi-crop            & 3096 & 43.4          & 67.8          & 5.4           & 51.8 \\
Multi-crop with S$^2$ & 2048 & 43.4          & \textbf{79.1} & \textbf{10.6} & \textbf{57.1} \\
Dynamic resolution    & 2048 & \textbf{45.2} & \textbf{81.5} & 9.2           & 51.3 \\
Dynamic resolution    & 3600 & \textbf{44.9} & 79.7          & \textbf{17.5} & \textbf{56.0} \\
\bottomrule
\end{tabular}
\end{table}

We trained a smaller (5B) variation of our model on a dataset of 10M image-text pairs, primarily composed of computer-use and GUI grounding data and experimented with several vision encoder configurations:
\begin{itemize}[leftmargin=1.5em, itemsep=2pt]
    \item \textbf{Dynamic S$^2$}~\citep{liu2025}: similar to S$^2$, but resizes to a rectangular resolution chosen to minimize distortion while admitting a tiling by $384 \times 384$ squares.
    \item \textbf{Multi-crop}: crops the image into (potentially overlapping) $384 \times 384$ squares; sends each through the vision encoder and concatenates features on the token dimension.
    \item \textbf{Multi-crop with S$^2$}: similar to multi-crop but uses S$^2$ to broaden the receptive field, i.e., crops the image into (potentially overlapping) $1536 \times 1536$ squares, performs S$^2$, and concatenates features on the token dimension.
    \item \textbf{Dynamic resolution}: a natively dynamic resolution vision encoder; we primarily used the NaFlex variant of the SigLIP-2 encoder~\citep{siglip2} and adjusted the minimum and maximum number of patches.
\end{itemize}

Our primary finding is that dynamic resolution vision encoders with a large number of visual tokens perform uniformly well, and the best on high-resolution datasets. It is particularly interesting to compare dynamic resolution with 2048 vs.\ 3600 maximum tokens: the latter roughly corresponds to native HD 720p resolution and enjoys a substantial boost on high-resolution benchmarks, particularly ScreenSpot-Pro. Reinforcing the high-resolution trend, we find that multi-crop with S$^2$ outperforms standard multi-crop despite using fewer visual tokens (i.e., fewer crops overall). Finally, it is worth noting that the dynamic resolution technique produces the most tokens on average; due to their tiling subroutine, S$^2$-based methods are constrained by the original image resolution and often only use about half the maximum tokens.

\paragraph{Open research questions.} While an increase in the resolution of the vision encoder substantially improves performance on high-resolution reasoning tasks, it comes at the cost of efficiency due to the quadratic complexity of attention with respect to the context length. With that said, each featurization technique we tested operates independently of the text prompt. It is an open question how to leverage text-conditioning to most efficiently tokenize the image---for example, if a specific question is asked about a high-resolution scene, the background could be encoded in a lower resolution to save on tokens. Similar ideas are present in the literature (e.g., the Q-Former from BLIP-2~\citep{blip2}), but their initial promise has not yet been proven out for agentic tasks.

\begin{table}[t!]
\centering
\caption{Training recipe for \phivr. Trainable modules are indicated with \checkmark; frozen modules with \texttimes.}
\label{tab:training-recipe}
\tablestyle{5pt}{1.25}
\begin{tabular}{l c c c l}
\toprule
\textbf{Stage} & \textbf{MLP} & \textbf{Vision Encoder} & \textbf{LLM} & \textbf{Data} \\
\midrule
1.\;MLP Pretraining          & \checkmark & \texttimes & \texttimes & Image--text alignment \\
2.\;Instruction Tuning       & \checkmark & \checkmark & \checkmark & Single-image instruction tuning \\
3.\;Long Context \& RAI      & \checkmark & \checkmark & \checkmark & Long content, multi-image, RAI \\
\bottomrule
\end{tabular}
\end{table}

\begin{table}[b!]
\centering
\caption{Training hyperparameters by stage.}
\label{tab:training-hparams}
\tablestyle{5pt}{1.25}
\begin{tabular}{l c c c}
\toprule
\textbf{Hyperparameter} & \textbf{Stage 1} & \textbf{Stage 2} & \textbf{Stage 3} \\
\midrule
Learning rate            & $1 \times 10^{-3}$ & $2 \times 10^{-5}$ & $7 \times 10^{-7}$ \\
LR schedule              & Cosine              & Cosine w/ min LR    & Cosine w/ min LR \\
Min LR ratio             & --                  & 0.1                 & 0.1 \\
Warmup                   & 3\% of steps        & 500 steps           & 500 steps \\
Weight decay             & 0                   & $10^{-4}$           & $10^{-4}$ \\
Adam $(\beta_1, \beta_2)$ & (0.9, 0.999)       & (0.9, 0.95)         & (0.9, 0.95) \\
Max grad norm            & 1.0                 & 1.0                 & 1.0 \\
Global batch size        & 1024                & 1920                & 1920 \\
Max sequence length      & 2048                & 8192                & 16384 \\
Training samples         & 2.0M                & 62.8M               & 3.2M \\
Training tokens          & 1.4B                & 188.5B              & 12B \\
\bottomrule
\end{tabular}
\end{table}

\subsection{Training Recipe}
\label{sec:training-recipe}

\phivr was trained in three stages, summarized in \tabref{tab:training-recipe}. The 1st stage trains the MLP only, with the rest of the model frozen, to warm-up the MLP from its random initialization.  This stage is relatively light and uses a small amount of clean image-captioning data to create initial alignment between the vision-encoder and LLM backbone.  We experimented with larger amounts of pretraining in this initial stage and saw no benefit. Stage 2 is the bulk of the training and trains the whole model unfrozen on a larger amount of single-image visual instruction tuning data.  Stage 3 is a medium sized stage that focuses on long-context, multi-image, and safety (RAI) training. \tabref{tab:training-hparams} lists the key optimization hyperparameters and training sample and token counts for each stage. All stages use the AdamW optimizer, bf16 mixed precision, DeepSpeed ZeRO-1, and train for one epoch.  More details on each stage:

\paragraph{Stage~1: MLP Pretraining.} Only the cross-modality projector (MLP) is trained while the vision encoder and language model remain frozen. This stage aligns the visual feature space of SigLIP-2 with the text embedding space of Phi-4-Reasoning, establishing a shared representation before any other parameters are updated.

\paragraph{Stage~2: Instruction Tuning.} All model components---the MLP, vision encoder, and language model---are jointly trained on single-image instruction-tuning data. This stage constitutes the bulk of training and covers the full range of tasks: visual question answering, mathematical and scientific reasoning, grounding, captioning, OCR, and computer-use. The mixture includes both reasoning traces (with \texttt{<think>} tokens) and direct-response samples (with \texttt{<nothink>} tokens) as described in \secref{sec:reasoning}.

\paragraph{Stage~3: Long Context, Multi-Image, and RAI.} The full model continues training on specialized data: long-document understanding, multi-image and sequential-image tasks, and additional responsible AI (RAI) data. This stage extends the model's capabilities to handle longer contexts and multi-turn visual interactions while reinforcing safety alignment.

\tabref{tab:data-sources} in the Appendix lists the public training data sources used across all stages, grouped by category. The majority of our data originates from public open-source datasets which were filtered and improved as described in \secref{sec:data}.
\section{Training Data}
\label{sec:data}

As with its language backbone Phi-4-Reasoning~\citep{phi4reasoning}, \phivr was trained with a deliberate focus on data quality. Our final data mix consists of data primarily from three sources: open-source vision-language datasets which were meticulously filtered and improved, high-quality domain-specific data from other Microsoft teams, and high-quality data from targeted acquisitions. The overwhelming majority of our data lies in the first category: data which originated as open-source data, after which a significant amount of effort was dedicated to filtering and improving, whether by removing low-quality datasets or records, programmatically fixing errors in data formatting, or using open-source images as seeds to synthetically generate higher-quality accompanying text.

\begin{figure}[t]
\centering
\includegraphics[width=0.99\linewidth]{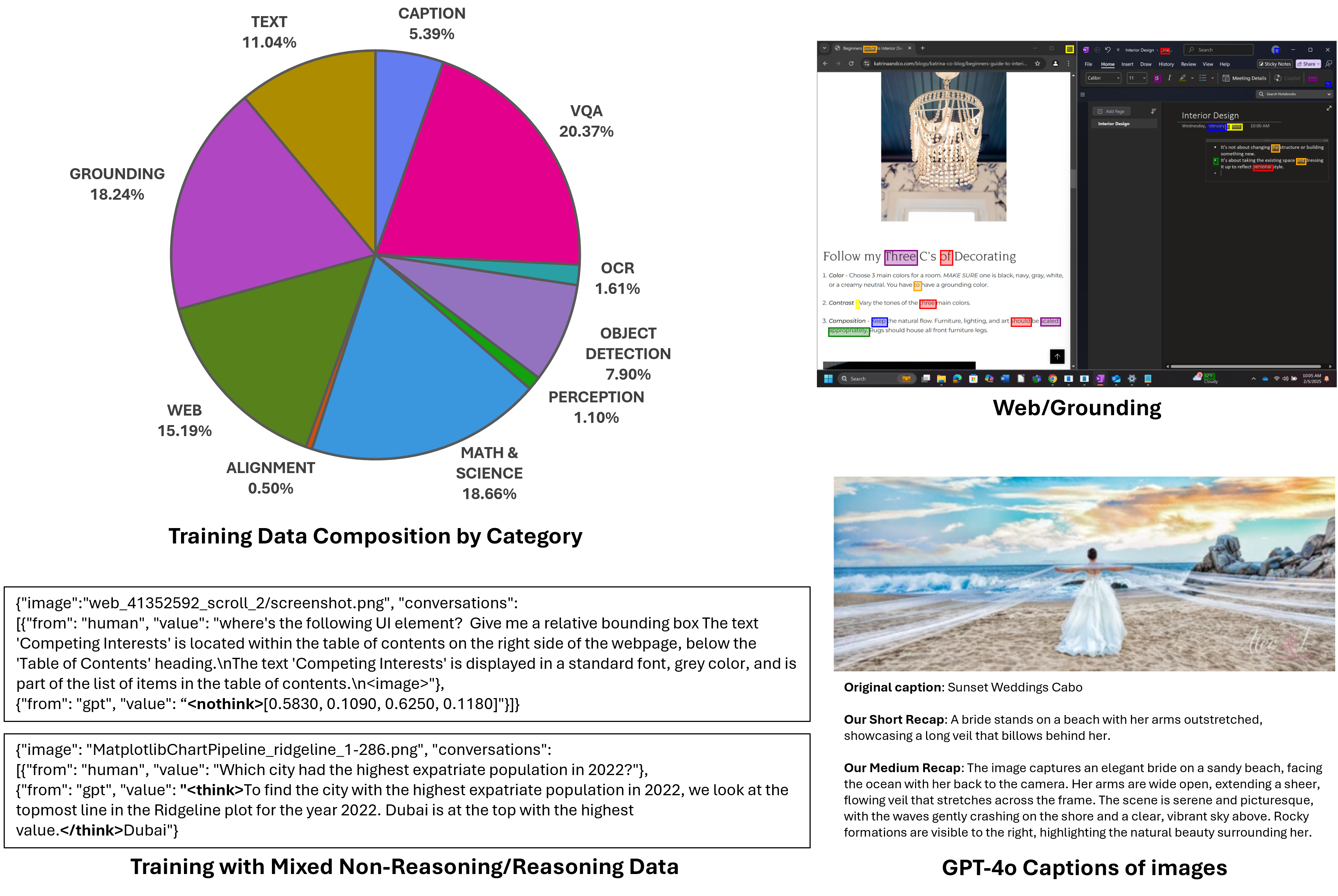}
\caption{Training data composition and examples for the Stage 2 training of \phivr.  The Stage 3 data is designed to have a similar composition.}
\label{fig:data-composition}
\end{figure}

\subsection{Data Quality}
\label{sec:dataquality}

The process of improving open-source data began by simply spending time manually sifting through data. Going through samples from each dataset, we found that 5--10 minutes per dataset was enough to classify in one or more of the following categories:

\begin{itemize}[leftmargin=1.5em, itemsep=2pt]
    \item \textbf{Excellent-quality data}: the text components of the data consist of high-quality questions paired with correct answers. The threshold for ``excellent'' data is somewhat category dependent; for example, high-quality caption data might look different from high-quality chart QA data.
    \item \textbf{Good questions with wrong answers}: the text components of the data consist of high-quality questions, answerable from the image, with some portion of incorrect answers. This category arises most commonly with diagram/math/STEM QA.
    \item \textbf{Low-quality questions}: the text components of the data contain some number of low-quality questions, which are either nonsensical or not answerable from the given image.
    \item \textbf{Low-quality images}: the images themselves are too repetitive or have fundamental errors (for example, a synthetic dataset of \LaTeX{} diagrams where text and figures tend to overlap chaotically).
    \item \textbf{High-quality with formatting errors}: the text component of the data contains formatting errors for many records, probably introduced during some processing stage; for example: all answers in a different format than what the prompt requests, misspelled image tags, final answers contained within reasoning blocks, etc.
\end{itemize}

Excellent-quality data was kept mostly unchanged, except perhaps for minor formatting improvements. For data with wrong answers or poor-quality captions, we re-generated answers or captions using GPT-4o and o4-mini, and, where appropriate, used the same models in verification or majority-voting pipelines. Not all such attempts succeeded, and we excluded a number of datasets with high percentages of wrong answers. We made some attempts to improve the data with low-quality questions, but we did not find much success with a naive approach (asking models for high-quality questions to complement images) except in some very special cases. However, when the images themselves were high quality, we used these as seeds to generate caption data or simple VQA, with the questions perhaps of a different flavor than the original text. We excluded datasets where the images themselves were fundamentally flawed. We fixed many formatting or logical errors, of which we found a surprising number across open-source datasets.

We employed a variety of techniques to get more mileage out of datasets, often with basic reformatting or diversification techniques, or using images as seeds to generate new image-text pairs. Some examples are as follows:

\begin{enumerate}[leftmargin=1.5em, itemsep=2pt]
    \item Based on the belief that Phi-4-Reasoning, as a language model backbone, can solve many VQA math problems provided that it can adequately interpret the mathematical elements of an image, we took every image from math/science/logic datasets and generated a detailed description of the image. This means that for all such domain-specific data, our data mix contains multiple records with the same image: one with the original QA and one with a caption-style description.
    \item Due to the limited amount of high-quality training data, we often asked our data to perform double-duty; for example, instead of having instruction-following data separate from the domain-specific data, we modified the text portion of data with ground-truth QA pairs to request and provide the answer in a specific format.
    \item After generating high-quality caption data from open-source images, we created multi-image data by creating records in scrambled and caption-matching formats. For the former, ${\sim}5$ images are given, and then captions are requested in a random order, and occasionally additional images are sprinkled in later. For the latter, ${\sim}5$ images are given, and the request is to match captions to images. We believe that such data improves the model's ability to attend to correct images in certain multi-image scenarios.
    \item With a similar goal to item 3, we generated ``what's changed?'' style data from pairs or triples of sequential screenshots, with the belief that such data improves the model's ability to better navigate images in real-time, as is necessary for CUA or robotics scenarios.
    \item We found that some datasets use overly complicated or over-engineered prompts (that is, the user half of the text portion) in their VQA data, which is likely to teach the model to only succeed in answering perfectly structured questions. We use a variety of human prompts to teach more robustness to the model.
\end{enumerate}

To supplement the improved open-source data, we utilize high-quality datasets shared with us by several teams at Microsoft, as well as several math-specific datasets which were acquired during training of the Phi-4 language model, and also some domain-specific curated data; for example, \LaTeX{}-OCR data generated by processing and rendering equations from arXiv documents.

\paragraph{Coordinate normalization.} Note that all spatial coordinates in our data are normalized to the range $[0.0, 1.0]$ relative to the image dimensions. This ensures a consistent representation across images of varying resolutions, and thus our model also always outputs normalized coordinates as well.

\subsection{Mathematics and Science vs.\ Computer-Use Data Proportion}
\label{sec:data-proportion}

One of our goals was to train a model which was simultaneously proficient at both mathematics and computer-use. It is an open question in the research community to understand how datasets should be structured to induce generalizable representations across diverse reasoning tasks. Importantly, how data scale affects reasoning performance can lead to starkly different design decisions, e.g., training a single model on a large dataset vs.\ multiple models with targeted post-training for mathematics and computer use.

Research on long-tailed classification robustness has suggested that balancing the data, or removing data from overrepresented tasks or subgroups, is an effective method for ensuring uniformly good performance~\citep{buda2018, idrissi2022, chaudhuri2023}. Nevertheless, these insights are at odds with the scale-is-all-you-need data paradigm. We conducted a suite of experiments to better understand optimal data scale and ratios for multimodal reasoning tasks of math and science reasoning vs. computer use -- our key focus areas for the model.

We trained a smaller variation of our model (5B parameters), 
while varying the amount of mathematics and computer-use data for each run. Each dataset included the same subset of 1M general image-text pairs as a baseline. For mathematics data, we used the same dataset of 150K multimodal records, optionally duplicating each one 3 times. Next, we included up to 450K computer-use records, and optionally an additional 400K from Phi-Ground~\citep{zhang2025}.

Our finding is that it appears possible for a single model to have uniformly superior performance across multiple reasoning domains. In general, multimodal mathematics performance was not harmed by additional computer-use data, and vice versa. The most impressive improvements were obtained on ScreenSpot-V2 by including the Phi-Ground dataset; its high specialization to GUI grounding reinforces the importance of targeted novel data collection. It is also worth noting that increasing mathematics data while keeping computer-use data constant still improves computer-use benchmarks.

\begin{table}[t]
\centering
\caption{Varying the ratios of math and CUA data. Increasing math data by 3$\times$ while keeping computer-use data constant improves both math and computer-use benchmarks.}
\label{tab:data-ratio}
\tablestyle{5pt}{1.2}
\begin{tabular}{c c c c c c c}
\toprule
\textbf{General} & \textbf{Math} & \textbf{CUA} & \textbf{Total} & \textbf{MMMU-CoT} & \textbf{MathVista} & \textbf{ScreenSpot-V2} \\
\midrule
1M & 150K & 450K & 1.6M & 44.0 & 37.4 & 48.2 \\
1M & 150K & 850K & 2.0M & 44.1 & 37.3 & 60.0 \\
1M & 450K & 450K & 1.9M & 45.3 & 36.0 & 48.3 \\
1M & 450K & 850K & 2.3M & 43.4 & 38.9 & \textbf{63.1} \\
1M & 150K & 150K & 1.3M & 44.2 & 36.9 & 29.8 \\
1M & 150K & 250K & 1.4M & \textbf{45.4} & 37.4 & 37.7 \\
\bottomrule
\end{tabular}
\end{table}

\paragraph{Open research questions.} Our experiments were conducted at a fairly small data scale wherein the model has not yet become saturated; in particular, overall performance correlated well with total data. A clear open question is to study the effects of data proportion at a scale which challenges the edge of current models' capabilities: do our insights about strong uniform performance hold, or do trade-offs between different reasoning tasks become more obvious at larger scales? Moreover, our imbalance ratios were fairly mild, with a 7.5\% ratio of mathematics data to total data at the worst. While well-studied in traditional machine learning settings such as long-tailed classification, understanding data dynamics at more extreme ratios (1\% or less) is an open problem, especially for performance on competing reasoning tasks.

\section{Mixed Non-Reasoning and Reasoning}
\label{sec:reasoning}

In language-only settings, reasoning traces have improved performance on many tasks, but they require additional compute which adds undesired latency. In multimodal settings, this tradeoff is less clear-cut: for tasks such as image captioning and optical character recognition (OCR), reasoning is often unnecessary and can even be harmful, while mathematical and scientific problem-solving benefit from multi-step reasoning. Thus, the choice of when to reason or not can be quite nuanced.

\subsection{Training Approaches for Multimodal Reasoning Models}

Language-only reasoning models are typically created through supervised fine-tuning (SFT) or reinforcement learning (RL): SFT is simpler but requires large amounts of expensive reasoning trace data, while RL reduces data requirements at the cost of significantly increased training complexity and compute. Multimodal reasoning models follow a similar process, but the design space is more complex. With a mid-fusion architecture, the first decision is whether the base language model is itself a reasoning or non-reasoning model. This leads to several possible training pipelines:

\begin{enumerate}[leftmargin=1.5em, itemsep=4pt]
    \item \textbf{Non-Reasoning LLM $\rightarrow$ Reasoning Multimodal Training:} Reasoning and multimodal capabilities are trained together.

    \item \textbf{Non-Reasoning LLM $\rightarrow$ Non-Reasoning Multimodal $\rightarrow$ Reasoning Multimodal Training:} Multimodal capabilities are learned first, then reasoning is added.

    \item \textbf{Reasoning LLM $\rightarrow$ Reasoning Multimodal Training:} A reasoning base is used, but all multimodal data must include reasoning traces.

    \item \textbf{Our approach: Reasoning LLM $\rightarrow$ Mixed Non-Reasoning / Reasoning Multimodal Training.} A reasoning-capable base is trained on a hybrid data mixture, learning when to reason and when to respond directly.
\end{enumerate}

Approaches 1 and 2 offer flexibility in designing multimodal reasoning behavior from scratch using widely available non-reasoning LLM checkpoints but place a heavy burden on multimodal training. Approach~1 must teach visual understanding and reasoning simultaneously and requires a large amount of multimodal reasoning data, while Approach~2 can be trained with less reasoning data but risks catastrophic forgetting, as reasoning training may degrade previously learned visual capabilities. Both risk weaker reasoning than starting from a reasoning-capable base. Approach~3 inherits strong reasoning foundations, but like Approach~1, it requires reasoning traces for all training data and produces reasoning traces for all queries, even when not beneficial.

\subsection{Our Approach: A Mixed Reasoning and Non-Reasoning Model}
\label{sec:our-approach}

\phivr adopts the 4\textsuperscript{th} approach listed previously, as it balances reasoning capability, inference efficiency, and data requirements. It inherits a strong reasoning foundation but uses a hybrid approach to combine the strengths of alternatives while mitigating their drawbacks. Our model defaults to direct inference for perception-focused domains where reasoning adds latency without improving accuracy, avoiding unnecessary verbosity and reducing inference costs, and it invokes longer reasoning paths for domains, such as math and science, that benefit from structured multi-step reasoning.

\paragraph{Implementation details.} Our model is trained with SFT, where reasoning samples include \texttt{<think>...</think>} sections with chain-of-thought reasoning before the final answer, covering domains like math and science. Non-reasoning samples are tagged to start with a \texttt{<nothink>} token, signaling a direct response, and cover perception-focused tasks such as captioning, grounding, OCR, and simple VQA. Reasoning data comprises approximately 20\% of the total mix. Starting from a reasoning-capable backbone means this data grounds existing reasoning in visual contexts rather than teaching it to reason from scratch.

\paragraph{Limitations and open questions.} This approach is not without limitations. The balance between modes is a direct function of design choices we made, informed by recent literature and observed model behavior during training. However, the boundary between modes can be imprecise as it is learned implicitly from the data distribution. Our model allows users to control this behavior through explicit prompting with \texttt{<think>} or \texttt{<nothink>} tokens when they want to override the default reasoning behavior. The 20/80 reasoning-to-non-reasoning data split may not be optimal for all domains or deployment contexts. Determining the ideal data balance, and ensuring that the model switches appropriately between modes, remains an open research problem.

We view this mixed approach not as a definitive solution, but as one well-motivated point in the design space for balancing latency, accuracy, and flexibility in multimodal systems.

\begin{figure}[t!]
    \centering
    \includegraphics[width=0.85\linewidth]{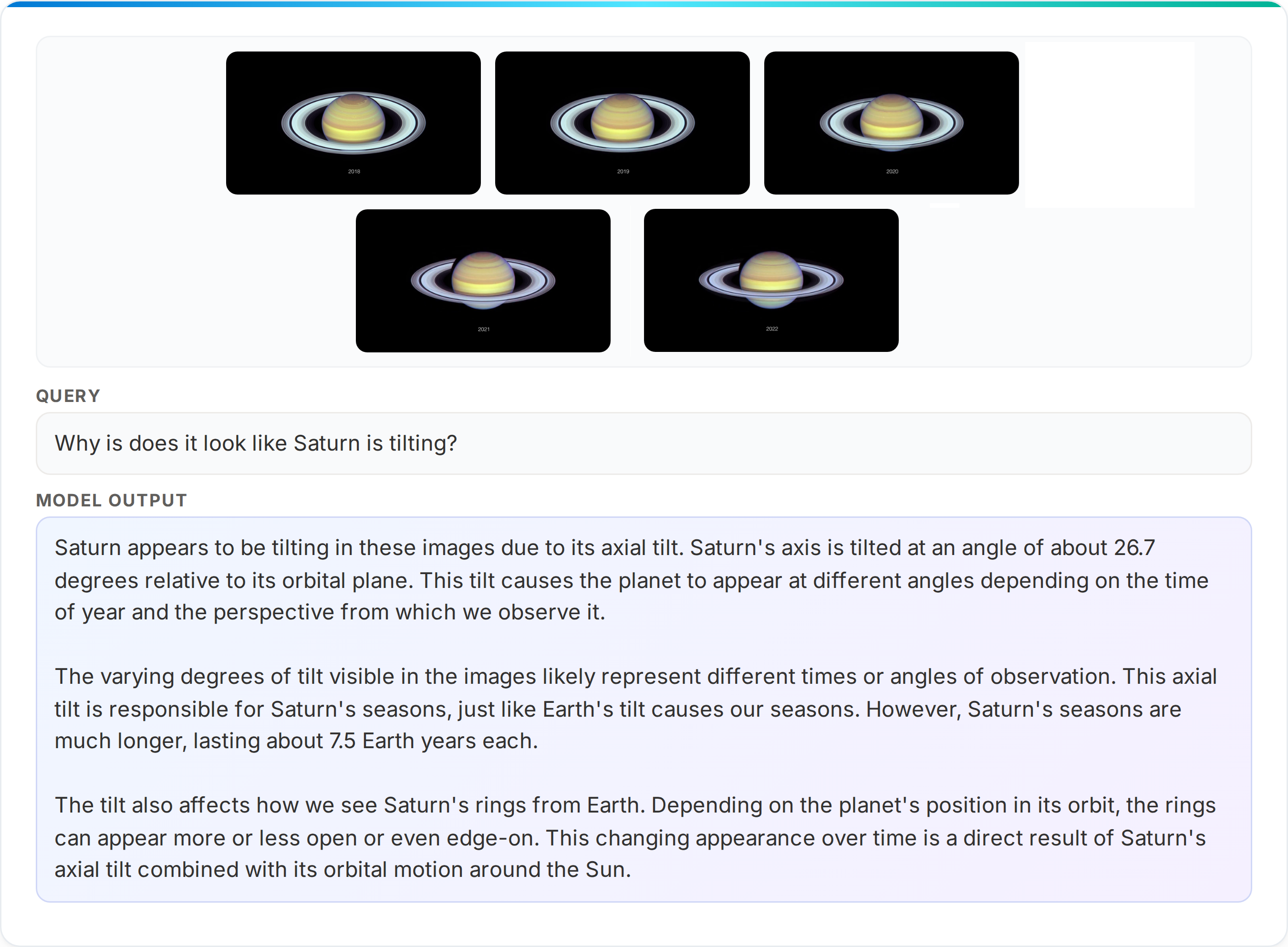}
    \caption{\phivr can interpret sequences of images, here reasoning about the changing appearance of Saturn's rings across multiple frames.}
    \label{fig:saturn}
\end{figure}

\section{Applications}
\label{sec:applications}

\phivr is a high-performing model across many vision-language tasks. It sees and understands the world by looking at a photo, document, chart, or screen and making sense of it. In practice that covers an enormous range of applications---just a few examples include: describing images and answering questions about them, interpreting changes and trends in image sequences, and recognizing objects, landmarks, and transcribing text.  Several examples are shown in Figure~\ref{fig:everyday} and ~\ref{fig:saturn}.

\subsection{Scientific and Mathematical Reasoning}

\begin{figure}[t!]
    \centering
    \includegraphics[width=0.85\linewidth]{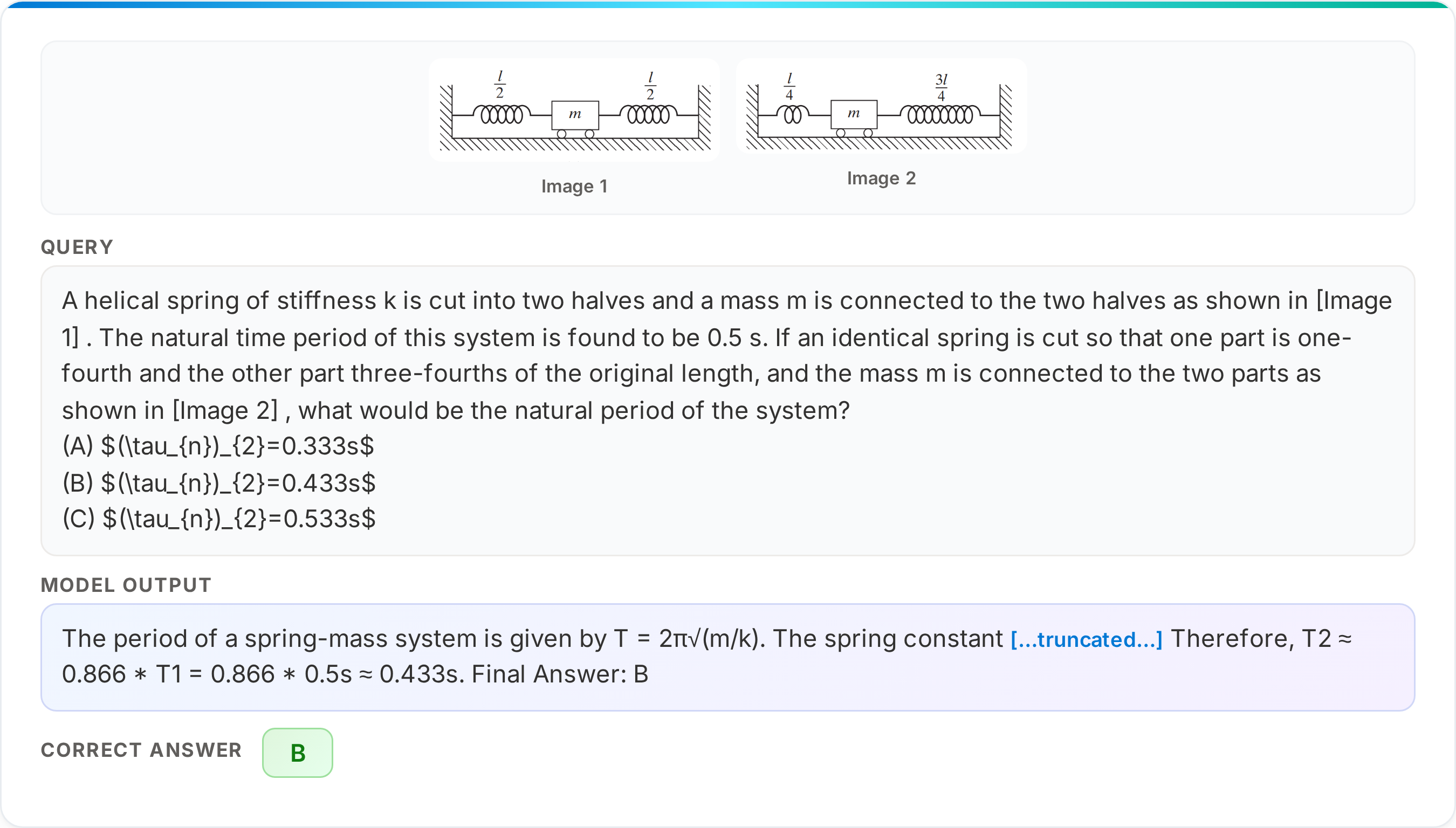}
    \caption{\phivr excels at math and science reasoning, correctly solving a multi-part spring-mass physics problem presented with diagrams.}
    \label{fig:math}
\end{figure}

\begin{figure}[t!]
    \centering
    \includegraphics[width=0.9\linewidth]{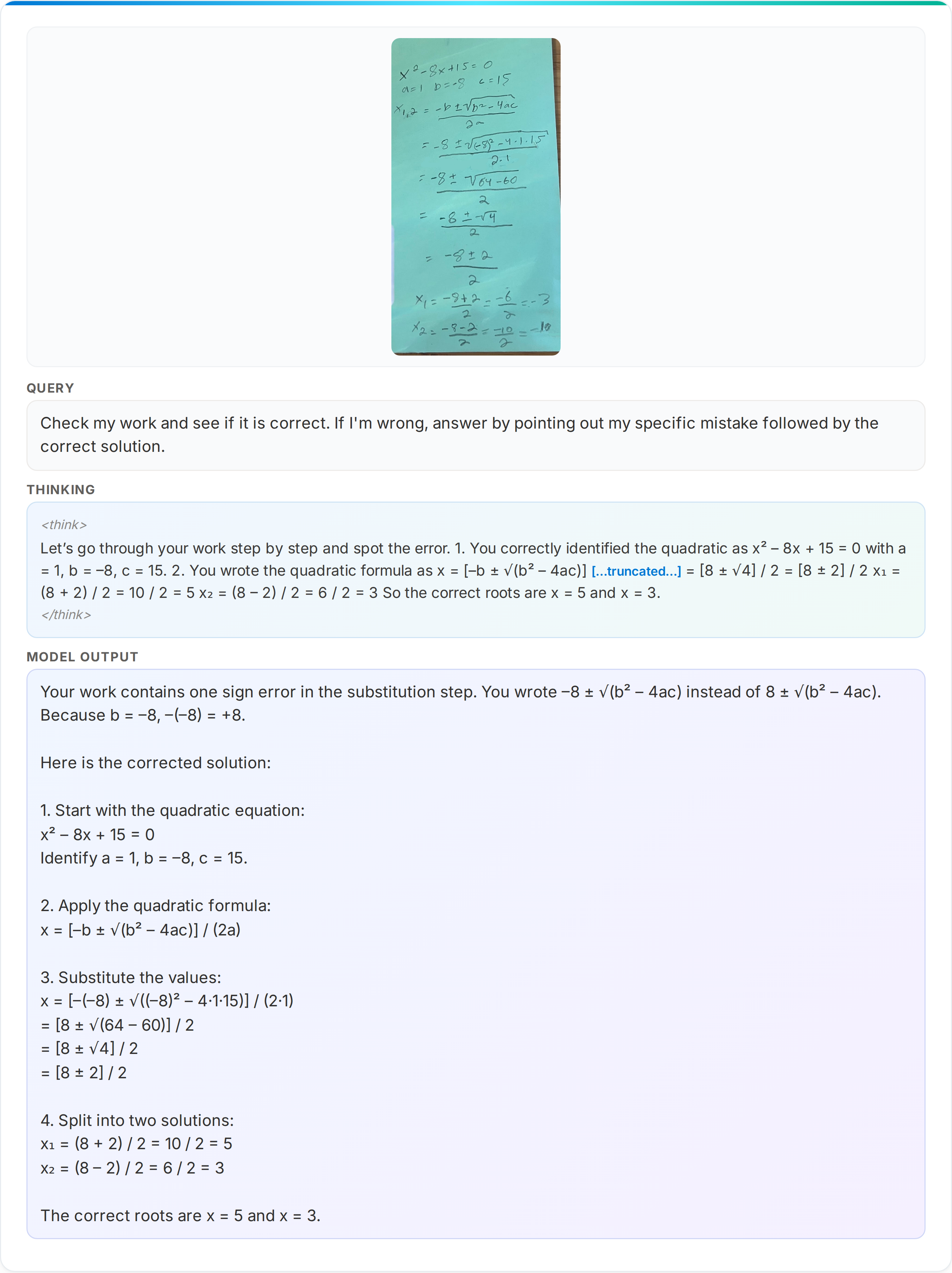}
    \caption{\phivr can help with written math problems, identifying a sign error in a handwritten quadratic formula solution and providing a corrected step-by-step derivation.}
    \label{fig:math-homework}
\end{figure}

In addition to general vision and language tasks, \phivr was designed to excel at tasks that combine visual input with structured inference, such as solving math problems presented in visual form (e.g., handwritten or diagram-based questions), extracting and reasoning over quantitative information in documents and charts, and supporting multi-step reasoning in educational or scientific analysis contexts.  Some examples are shown in Figure~\ref{fig:math}
and ~\ref{fig:math-homework}.

\subsection{Computer-Using Agents (CUA)}

\begin{figure}[t]
    \centering
    \begin{subfigure}[b]{0.48\linewidth}
        \centering
        \includegraphics[width=\linewidth]{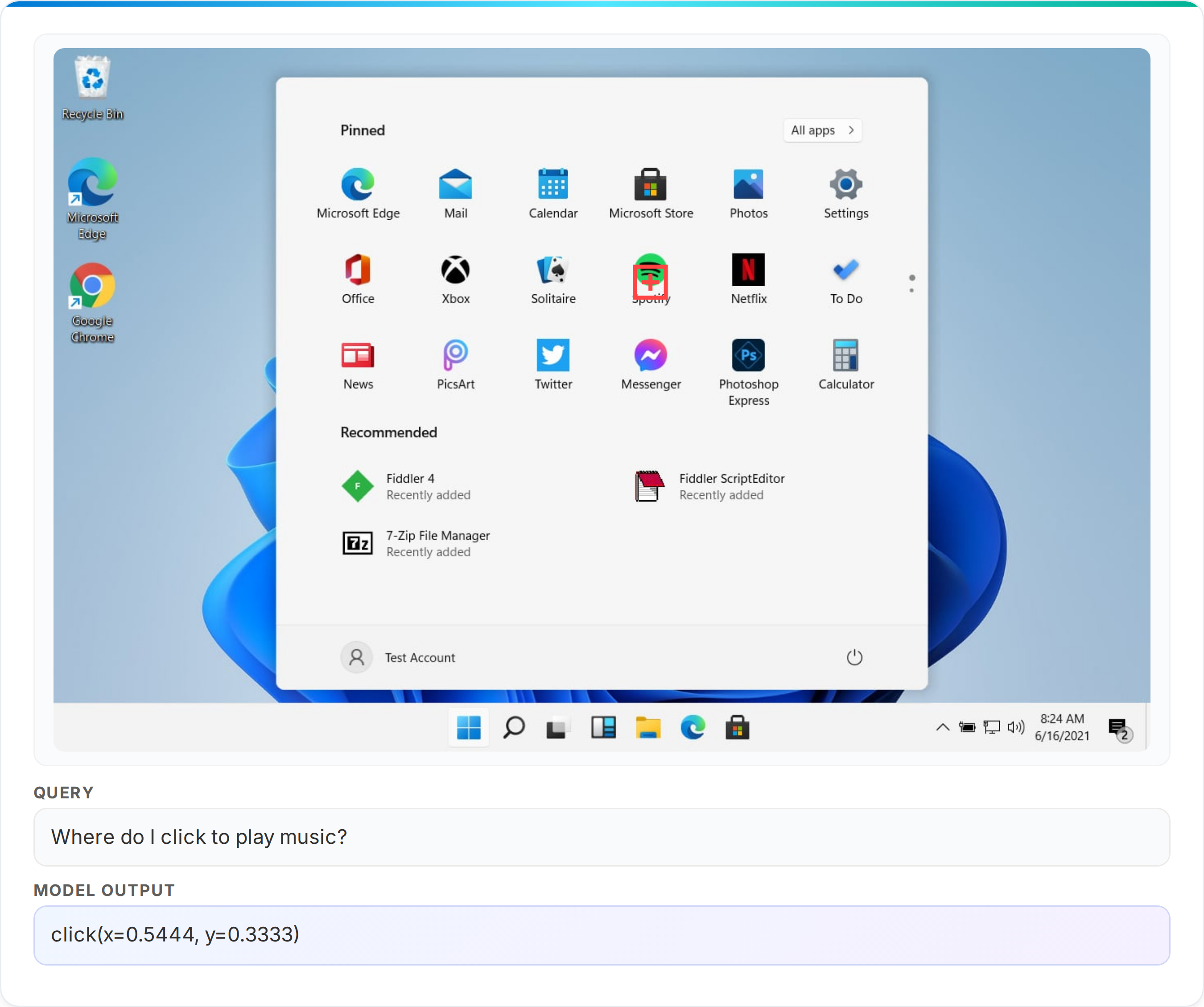}
        \caption{GUI grounding on a Windows desktop.}
    \end{subfigure}
    \hfill
    \begin{subfigure}[b]{0.48\linewidth}
        \centering
        \includegraphics[width=\linewidth]{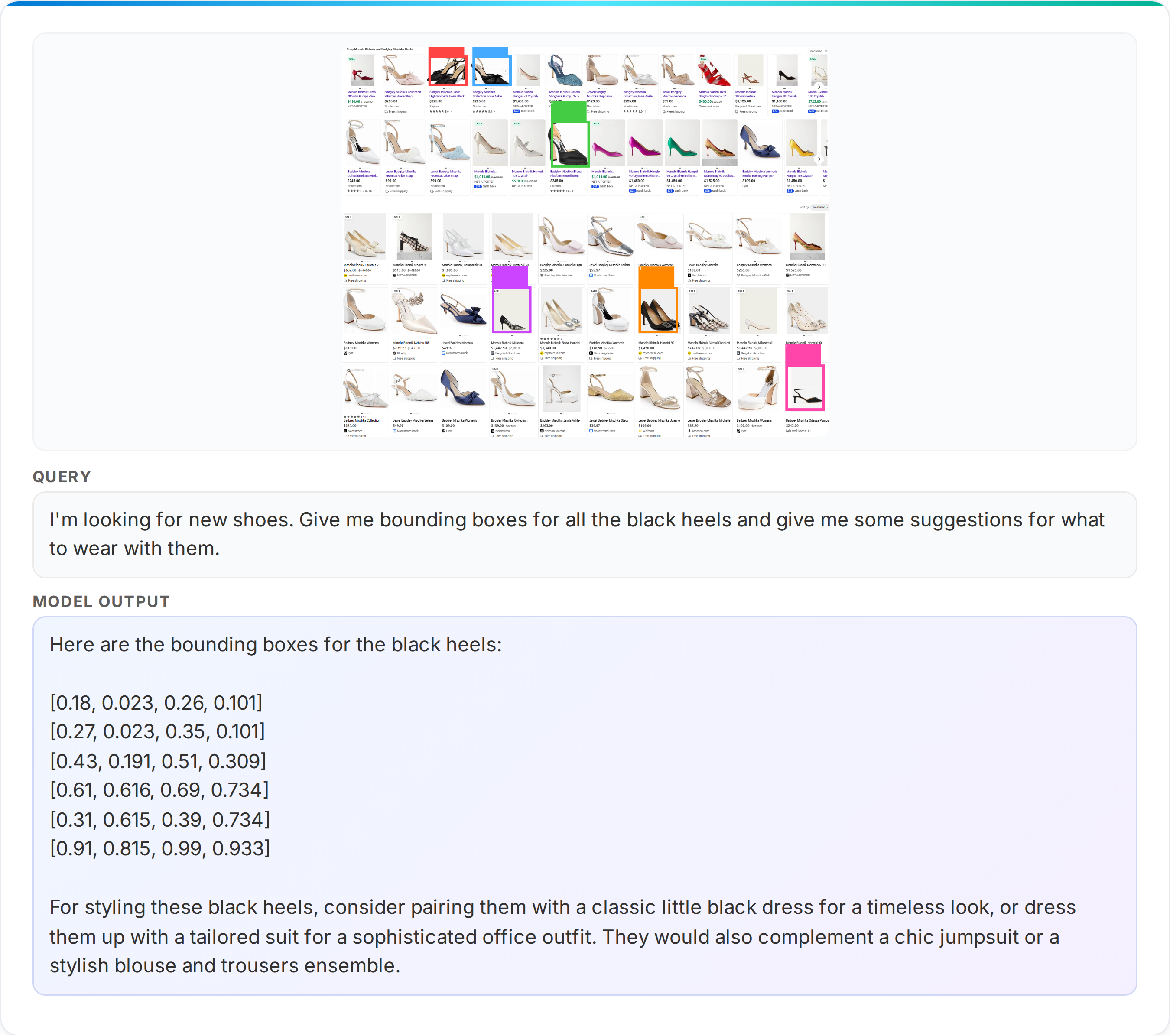}
        \caption{Object grounding in a product catalog.}
    \end{subfigure}
    \caption{\phivr can help navigate computer UIs, grounding interactive elements on desktop interfaces and localizing objects across dense visual layouts.}
    \label{fig:cua}
\end{figure}

We trained Phi-4-reasoning-vision-15B to develop capabilities that enable agents to interact with graphical user interfaces (GUIs). The model can interpret screen content and identify appropriate actions. Some examples are shown in Figure~\ref{fig:cua}.
With strong high-resolution perception and fine-grained grounding abilities, Phi-4-reasoning-vision-15B provides a strong foundation for building agentic systems.
These systems can navigate desktop, web, and mobile environments by detecting and localizing interactive elements such as buttons, menus, and text fields. The model’s visual understanding and spatial grounding allow it to reason about interface structure and determine appropriate interactions.
Due to its low inference-time needs, it is well-suited for interactive environments where low latency and compact model size are essential.

\section{Evaluation}
\label{sec:evaluation}

\phivr was evaluated for accuracy and timing using two complementary open-source frameworks to ensure both rigorous and standardized analysis: Eureka ML Insights\footnote{\url{https://github.com/microsoft/eureka-ml-insights}} and VLMEvalKit\footnote{\url{https://github.com/open-compass/VLMEvalKit}}. We report results on the following benchmarks: AI2D~\citep{ai2d}, ChartQA~\citep{chartqa}, HallusionBench~\citep{hallusionbench}, MathVerse~\citep{mathverse}, MathVision~\citep{mathvision}, MathVista~\citep{mathvista}, MMMU~\citep{mmmu}, MMStar~\citep{mmstar}, OCRBench~\citep{ocrbench}, and ScreenSpot\textsubscript{v2}sc~\citep{seeclick}.  Accruacy results are presented for our model and compared to several current, open-weight non-thinking and thinking models in Tables~\ref{tab:non-thinking} and~\ref{tab:thinking}, respectively.

\begin{table}[b!]
\centering
\caption{Accuracy comparisons relative to popular open-weight, non-thinking models.}
\label{tab:non-thinking}
\tablestyle{3pt}{1.15}
\resizebox{\textwidth}{!}{%
\begin{tabular}{l c c c c c c c c c}
\toprule
\textbf{Benchmark} &
\textbf{\makecell{\phivr}} &
\textbf{\makecell{Phi-4-\\reasoning-\\vision-15B\\--force\\nothink}} &
\textbf{\makecell{Phi-4-\\mm-\\instruct}} &
\textbf{\makecell{Kimi-VL-\\A3B-\\Instruct}} &
\textbf{\makecell{gemma-\\3-\\12b-it}} &
\textbf{\makecell{Qwen3-\\VL-8B-\\Instruct\\-4K}} &
\textbf{\makecell{Qwen3-\\VL-8B-\\Instruct\\-32K}} &
\textbf{\makecell{Qwen3-\\VL-32B-\\Instruct\\-4K}} &
\textbf{\makecell{Qwen3-\\VL-32B-\\Instruct\\-32K}} \\
\midrule
AI2D\textsubscript{TEST}       & 84.8 & 84.7 & 68.6 & 84.6 & 80.4 & 82.7 & 83   & 84.8 & 85   \\
ChartQA\textsubscript{TEST}    & 83.3 & 76.5 & 23.5 & 87   & 39   & 83.1 & 83.2 & 84.3 & 84   \\
HallusionBench                 & 64.4 & 63.1 & 56   & 65.2 & 65.3 & 73.5 & 74.1 & 74.4 & 74.9 \\
MathVerse\textsubscript{MINI}  & 44.9 & 43.8 & 32.4 & 41.7 & 29.8 & 54.5 & 57.4 & 64.2 & 64.2 \\
MathVision\textsubscript{MINI} & 36.2 & 34.2 & 20   & 28.3 & 31.9 & 45.7 & 50   & 54.3 & 60.5 \\
MathVista\textsubscript{MINI}  & 75.2 & 68.7 & 50.5 & 67.1 & 57.4 & 77.1 & 76.4 & 82.5 & 81.8 \\
MMMU\textsubscript{VAL}        & 54.3 & 52   & 42.3 & 52   & 50   & 60.7 & 64.6 & 68.6 & 70.6 \\
MMStar                         & 64.5 & 63.9 & 45.9 & 60   & 59.4 & 68.9 & 69.9 & 73.7 & 74.3 \\
OCRBench                       & 76   & 75.6 & 62.6 & 86.5 & 75.3 & 89.2 & 90   & 88.5 & 88.5 \\
ScreenSpot\textsubscript{v2}   & 88.2 & 88.3 & 28.5 & 89.8 & 3.5  & 91.5 & 91.5 & 93.7 & 93.9 \\
\bottomrule
\end{tabular}%
}
\end{table}

\begin{table}[t!]
\centering
\caption{Accuracy comparisons relative to popular open-weight, thinking models.}
\label{tab:thinking}
\tablestyle{3pt}{1.15}
\resizebox{\textwidth}{!}{%
\begin{tabular}{l c c c c c c c c}
\toprule
\textbf{Benchmark} &
\textbf{\makecell{\phivr}} &
\textbf{\makecell{Phi-4-\\reasoning-\\vision-15B\\--force\\thinking}} &
\textbf{\makecell{Kimi-VL-\\A3B-\\Thinking}} &
\textbf{\makecell{gemma-\\3-12b-it}} &
\textbf{\makecell{Qwen3-\\VL-8B-\\Thinking\\-4K}} &
\textbf{\makecell{Qwen3-\\VL-8B-\\Thinking\\-40K}} &
\textbf{\makecell{Qwen3-\\VL-32B-\\Thinking\\-4K}} &
\textbf{\makecell{Qwen3-\\VL-32B-\\Thinking\\-40K}} \\
\midrule
AI2D\textsubscript{TEST}       & 84.8 & 79.7 & 81.2 & 80.4 & 83.5 & 83.9 & 86.9 & 87.2 \\
ChartQA\textsubscript{TEST}    & 83.3 & 82.9 & 73.3 & 39   & 78   & 78.6 & 78.5 & 79.1 \\
HallusionBench                 & 64.4 & 63.9 & 70.6 & 65.3 & 71.6 & 73   & 76.4 & 76.6 \\
MathVerse\textsubscript{MINI}  & 44.9 & 53.1 & 61   & 29.8 & 67.3 & 73.3 & 78.3 & 78.2 \\
MathVision\textsubscript{MINI} & 36.2 & 36.2 & 50.3 & 31.9 & 43.1 & 50.7 & 60.9 & 58.6 \\
MathVista\textsubscript{MINI}  & 75.2 & 74.1 & 78.6 & 57.4 & 77.7 & 79.5 & 83.9 & 83.8 \\
MMMU\textsubscript{VAL}        & 54.3 & 55   & 60.2 & 50   & 59.3 & 65.3 & 72   & 72.2 \\
MMStar                         & 64.5 & 63.9 & 69.6 & 59.4 & 69.3 & 72.3 & 75.5 & 75.7 \\
OCRBench                       & 76   & 73.7 & 79.9 & 75.3 & 81.2 & 82   & 83.7 & 85   \\
ScreenSpot\textsubscript{v2}   & 88.2 & 88.1 & 81.8 & 3.5  & 93.3 & 92.7 & 83.1 & 83.1 \\
\bottomrule
\end{tabular}%
}
\end{table}

As shown in Tables~\ref{tab:non-thinking} and~\ref{tab:thinking}, our model balances thinking and non-thinking performance---on average showing better accuracy in the default mixed-reasoning behavior than when forcing thinking vs.\ non-thinking. Only in a few cases does forcing a specific mode improve performance (MathVerse and MMMU\textsubscript{VAL} for thinking and ScreenSpot\textsubscript{v2} for non-thinking). 

\paragraph{Timing experiments.} To produce the accuracy-vs-compute plots in \figref{fig:timing-and-tokens}, we randomly sampled 100 examples from each of four benchmarks---ChartQA\textsubscript{TEST}, MathVista\textsubscript{MINI}, MMMU\textsubscript{VAL}, and ScreenSpot---and measured wall-clock latency and output token counts for every model. All timing experiments were conducted using Eureka ML Insights on NVIDIA H100 GPUs using a single thread with no concurrency and a batch size of one, in order to obtain the most accurate estimate of per-query latency similar to what a user would experience in an interactive setting. We initially tried with vLLM to performing timing experiments using the recommended parameters for each model; while this increased throughput, it also increased per-query latency, giving an inflated estimate of the interactive timing we aimed to measure.

Compared to recent popular, open-weight models, \phivr provides a desirable trade-off between accuracy and cost (as a function of inference time compute and output tokens).

\textbf{Note:} All numbers here are the result of running benchmarks ourselves and may be lower than other previously shared numbers. Instead of quoting leaderboards, we performed our own benchmarking, so we could understand scaling performance as a function of output token counts for related models. We made our best effort to run fair evaluations and used recommended evaluation platforms with model-specific recommended settings and prompts provided for all third-party models. For Qwen models we use the recommended token counts and also ran evaluations matching our max output token count of 4096. For \phivr, we used our system prompt and chat template but did not do any custom user-prompting or parameter tuning, and we ran all evaluations with temperature${}=0.0$, greedy decoding, and 4096 max output tokens. These numbers are provided for comparison and analysis rather than as leaderboard claims. For maximum transparency and fairness, we will release all our evaluation logs publicly.

\section{Safety}
\label{sec:safety}

As with other Phi models, \phivr was developed with safety as a core consideration throughout training and evaluation. The model was trained on a mixture of public safety datasets and internally generated examples designed to elicit behaviors the model should appropriately refuse, in alignment with Microsoft's Responsible AI Principles. These safety-focused training signals help the model recognize and decline requests that fall outside intended or acceptable use.

Specifically, Phase~3 of training (\secref{sec:training-recipe}) incorporates dedicated open-source, responsible AI (RAI) data, including Hateful Memes~\citep{hatefulmemes}, VLGuard~\citep{vlguard}, Think-in-Safety~\citep{thinkinsafety}, WildGuard~\citep{wildguard}. This data covers a range of safety-relevant scenarios such as hateful content detection, refusal of harmful requests, and safe reasoning under adversarial prompts.

Phi-4-Reasoning-Vision-15B's safety was evaluated using both quantitative and qualitative approaches. Automated red teaming was performed on Azure to assess safety risks across multiple risk categories, including disallowed content (sexual, violent, hateful, or self-harm content), copyright content and intellectual property, and jailbreak susceptibility. The evaluation assessed the model's groundedness and its tendency to generate fabricated or misleading information.
The safety evaluation built upon the established practices from the Phi-4-Reasoning model's safety assessment. The multimodal nature of the model introduces additional safety considerations around visual content interpretation, and evaluations were conducted to assess the model's behavior when presented with potentially harmful or misleading visual inputs.

\begin{table}[h]
\centering
\begin{tabular}{|l|l|r|}
\hline
\textbf{Evaluation} & \textbf{Description} & \textbf{Defect Rate} \\
\hline
Text to Text Safety & Automated content safety evaluation measuring safety policies & 1.4\% \\
\hline
Image to Text Safety & Automated content safety evaluation measuring safety policies & 4.5\% \\
\hline
\end{tabular}
\end{table}

\section{Limitations}
\label{sec:limitations}

While \phivr achieves strong results relative to its size and compute budget, several limitations should be noted:

\begin{itemize}[leftmargin=1.5em, itemsep=2pt]
    \item Larger proprietary models outperform on broad, unconstrained vision--language benchmarks and generalist multimodal tasks. \phivr is competitive with open-weight models of similar size, and achieves state-of-the-art accuracy relative to training and inference-time compute and tokens---less compute and fewer tokens translates to less waiting and reduced cost.
    \item The learned switching between reasoning and non-reasoning modes is not always optimal. In some cases, the model may reason when a direct response would suffice, or respond directly when reasoning would be beneficial. Explicit prompting with \texttt{<think>} or \texttt{<nothink>} tokens can be used to override the default behavior when needed.
    \item Like many models of its size, \phivr has limitations particularly around extremely detailed or nuanced understanding of images. Users should verify critical outputs, especially for fine-grained visual details.
\end{itemize}

\section{Open Release and Community Engagement}
\label{sec:release}

\phivr is available on Microsoft Foundry and HuggingFace with additional examples and details on GitHub. For additional guidance on how to use our model properly and safely, please refer to our Model Card.

In line with our goal of supporting future AI development in the community, \phivr is released under a permissive license with model weights, fine-tuning code, and benchmark logs. We plan to release a portion of our training data in the coming months. We intend this release to complement existing work by providing concrete artifacts that help close gaps in understanding how compact multimodal reasoning models can be built and studied.

\section{Looking Forward}
\label{sec:looking-forward}

Smaller vision--language models with selective, task-aware reasoning offer one promising direction for making multimodal systems more practical and accessible. We present our model and its learnings to inform ongoing research in multimodal modeling, computer-using agents, and mathematical scientific reasoning.

We hope these details are useful to researchers exploring similar tradeoffs and invite critical evaluation, replication, and extension by the community.

\section*{Acknowledgements}

We thank Rachel Ward for her extensive work on data collection and curation. We thank the GenDatasets, PhiGround, SimCity, and Fara-7B efforts for invaluable training data. We thank Harkirat Behl, Mojan Javaheripi, and Suriya Gunasekar for providing us with Phi-4 checkpoints and guidance on training with Phi models. We additionally thank Sahaj Agarwal, Ahmed Awadallah, Qi Dai, Gustavo de Rosa, Rafah Hosn, Ece Kamar, Piero Kauffmann, Yash Lara, Chong Luo, Caio César Teodoro Mendes, Akshay Nambi, Craig Presti, Matthew Rosoff, Corby Rosset, Marco Rossi, Kashyap Patel, Adil Salim, Sidhartha Sen, Shital Shah, Pratyusha Sharma, Alexey Taymanov, Vibhav Vineet, John Weiss, Spencer Whitehead, the AI Frontiers Team and Leadership, and Microsoft Research Leadership, for their valuable help, insightful discussions, and continued support throughout this work.

\bibliographystyle{plainnat}
\bibliography{main}

\newpage
\appendix

\section{Open-Source Training Data}
\label{sec:appendix-data}

\begin{table}[h]
\centering
\caption{Open-Source Training Data Sources for Stages~1--3.}
\label{tab:data-sources}
\tablestyle{4pt}{1.15}
\resizebox{\textwidth}{!}{%
\begin{tabular}{l l l}
\toprule
\textbf{Stage} & \textbf{Category} & \textbf{Datasets} \\
\midrule
\textbf{Stage 1: MLP Pretraining}
 & Image--Text Alignment & Bunny~\citep{bunny} \\
\midrule
\multirow{14}{*}{{\textbf{Stage 2: Single-Image Instruction Tuning}}}
 & Caption              & Bunny~\citep{bunny} Recaptioned, Pixmo~\citep{pixmo}, LLaVA-OneVision~\citep{llavanext} \\
 & Diagram \& Chart QA  & LLaVA-OneVision~\citep{llavanext}, CoSyn~\citep{cosyn} \\
 & Document QA          & Docmatix~\citep{docmatix}, LLaVA-OneVision~\citep{llavanext} \\
 & Object Detection     & Open Images~\citep{openimages} \\
 & OCR                  & LLaVA-OneVision~\citep{llavanext}, IAM~\citep{iam} \\
 & Perception           & LLaVA-OneVision~\citep{llavanext}, VisOnlyQA~\citep{visonlyqa} \\
 & Text                 & NuminaMath~\citep{numinamath}, OpenThoughts~\citep{openthoughts} \\
 & VQA                  & Bunny~\citep{bunny}, LLaVA-OneVision~\citep{llavanext}, LLaVA-NeXT~\citep{llavanext}, ShareGPT4V~\citep{sharegpt4v}, Pixmo~\citep{pixmo} \\
 & Math: OCR            & NuminaMath~\citep{numinamath} \\
 & Math: Problem        & LLaVA-OneVision~\citep{llavanext}, NuminaMath~\citep{numinamath}, MMR1~\citep{mmr1}, Eedi~\citep{eedi} \\
 & CUA: General         & AGUVis~\citep{aguvis}, MultiUI~\citep{multiui}, Pixmo~\citep{pixmo}, CoSyn~\citep{cosyn}, BigDocs~\citep{bigdocs} \\
 & CUA: Grounding       & PhiGround~\citep{zhang2025}, SeeClick~\citep{seeclick} \\
 & CUA: HTML            & WebSight~\citep{websight} \\
 & RAI                  & Hateful Memes~\citep{hatefulmemes}, Think-in-Safety~\citep{thinkinsafety}, WildGuard~\citep{wildguard} \\
\midrule
\multirow{6}{*}{{\textbf{Stage 3: Long Context, Mulit-Image, and RAI}}}
 & Caption              & Bunny~\citep{bunny} \\
 & Document QA          & Docmatix~\citep{docmatix} \\
 & VQA                  & M4-Instruct~\citep{m4instruct} \\
 & Math \& Science      & VisualWebInstruct~\citep{visualwebinstruct} \\
 & CUA                  & AGUVis~\citep{aguvis} \\
 & RAI                  & Hateful Memes~\citep{hatefulmemes}, VLGuard~\citep{vlguard}, Think-in-Safety~\citep{thinkinsafety}, WildGuard~\citep{wildguard} \\
\bottomrule
\end{tabular}%
}
\end{table}

\end{document}